\documentclass[10pt,twocolumn,letterpaper]{article}

\usepackage{cvpr}      

\usepackage{microtype}

\renewcommand{\paragraph}[1]{\vspace{.5em}\noindent\textbf{#1.}}

\usepackage{soul}
\setuldepth{foobar}

\definecolor{cvprblue}{rgb}{0.21,0.49,0.74}
\usepackage[pagebackref,breaklinks,colorlinks,allcolors=cvprblue]{hyperref}

\usepackage[utf8]{inputenc} 
\usepackage[T1]{fontenc}    
\usepackage{url}            
\usepackage{booktabs}       
\usepackage{amsfonts}       
\usepackage{nicefrac}       
\usepackage{microtype}      
\usepackage{xcolor}         
\usepackage{multirow}  
\usepackage{graphicx}
\usepackage{amsmath}
\usepackage{wrapfig}

\definecolor{cvprblue}{rgb}{0.21,0.49,0.74}
\usepackage[pagebackref,breaklinks,colorlinks,allcolors=cvprblue]{hyperref}
\newcommand{\approachName}[1]{\textit{Video-GMAE}} 

\def\paperID{*****} 
\def\confName{CVPR}
\def\confYear{2026}

\title{Tracking by Predicting 3-D Gaussians Over Time}

\author{
Tanish Baranwal \and Himanshu Gaurav Singh \and Jathushan Rajasegaran \and Jitendra Malik \\
University of California, Berkeley \\
}
\begin{document}

\maketitle

\begin{abstract}
We propose Video Gaussian Masked Autoencoders (\approachName{x}), a self-supervised approach for representation learning that encodes a sequence of images into a set of Gaussian splats moving over time. Representing a video as a set of Gaussians enforces a reasonable inductive bias: that 2-D videos are often consistent projections of a dynamic 3-D scene. We find that tracking emerges when pretraining a network with this architecture. Mapping the trajectory of the learnt Gaussians onto the image plane gives zero-shot tracking performance comparable to state-of-the-art. With small-scale finetuning, our models achieve 34.6\% improvement on Kinetics, and 13.1\% on Kubric datasets, surpassing existing self-supervised video approaches.
\begingroup
\renewcommand\thefootnote{}
\footnotetext{Project page: \url{https://videogmae.org/}. Code: \url{https://github.com/tekotan/video-gmae}.}
\endgroup


 
\end{abstract}

\vspace{-0.4cm}
\section{Introduction}


Each pixel's journey through a video tells a story about motion. By tracking such pixels, we can understand the structure of the scene and interaction among its constituents. Smooth-pursuit tracking emerges in children as a fundamental visual skill in the first 2-4 months of their lives, before acquiring 3-D understanding, sensorimotor control, and object permanence~\cite{piaget21954construction}. Similarly, in computer vision systems, solving correspondence is essential for computational photography, 3-D understanding, and other long-range reasoning tasks. While pixel tracking has been studied in depth in vision literature~\cite{doersch2022tap, le2024dense, karaev2024cotracker, ravi2024sam}, it has been limited to training models on annotated point tracks, bounding boxes, segmentation masks~\cite{karaev2024cotracker, dendorfer2021motchallenge, ravi2024sam}, synthetic datasets~\cite{teed2020raft} or specialized architectures~\cite{jabri2020space, shrivastava2024self, stojanov2025self}. In this paper, we present a self-supervised method on videos that learns strong representations for correspondence.

We find that the representations learned by existing video self-supervised learning (SSL) approaches do not perform well on point tracking. We hypothesize that the classic (space-)time patch prediction objective does not strongly enforce temporal consistency; that is, this objective can be optimized without understanding pixel-level correspondence across a long sequence of frames. 

How can we set up our SSL task such that this correspondence emerges in the representations? We note that the motion of objects in 3-D manifests as point tracking on the image plane. In this paper, we leverage this insight to learn point tracking by pre-training on unlabeled videos. 

We train a Masked Autoencoder~\cite{he2022masked}-style encoder-decoder architecture that takes video as input and predicts \textit{Gaussian primitives} \cite{Kerbl2023} that move in time. We call our approach \approachName{}, in which for the first frame, we predict a fixed number of Gaussian primitives. For subsequent frames, we predict the \textit{translation} and \textit{color} change of each Gaussian with respect to the previous frame. In this way, each Gaussian preserves its identity over the span of N frames. Encoding a frame sequence as a set of moving Gaussians explicitly imposes temporal correspondence in 3-D as an inductive bias in our training. This inductive bias makes the SSL task \textit{harder}, inducing the latent representations to encode long-term correspondence.  

To investigate our pretraining methodology, we devise an algorithm to compute point tracks from the predicted 4-D Gaussians \textit{zero-shot}. We find that a zero-shot tracker naturally emerges from our correspondence-aware Gaussian representations. \approachName{} also gives latent representations that are useful for tracking. Finetuning \approachName{} on tracking datasets outperforms prior supervised and self-supervised methods. \approachName{} also outperforms other video SSL approaches such as VideoMAE~\cite{Tong2022}, MAE-ST~\cite{Feichtenhofer2022} on frozen encoder tracking evaluation. 

Broadly, we aim to bring classic end-to-end SSL together with differentiable rendering. Differentiable rendering is a natural inductive bias for video SSL: since videos are basically 2-D projections of a 3-D world, dynamic yet consistent over time. We hope that this work aids the search for video-SSL methods that learn longer-horizon and more generalizable representations.

\begin{figure*}[!tb]
    \centering
    \includegraphics[width=0.95\linewidth]{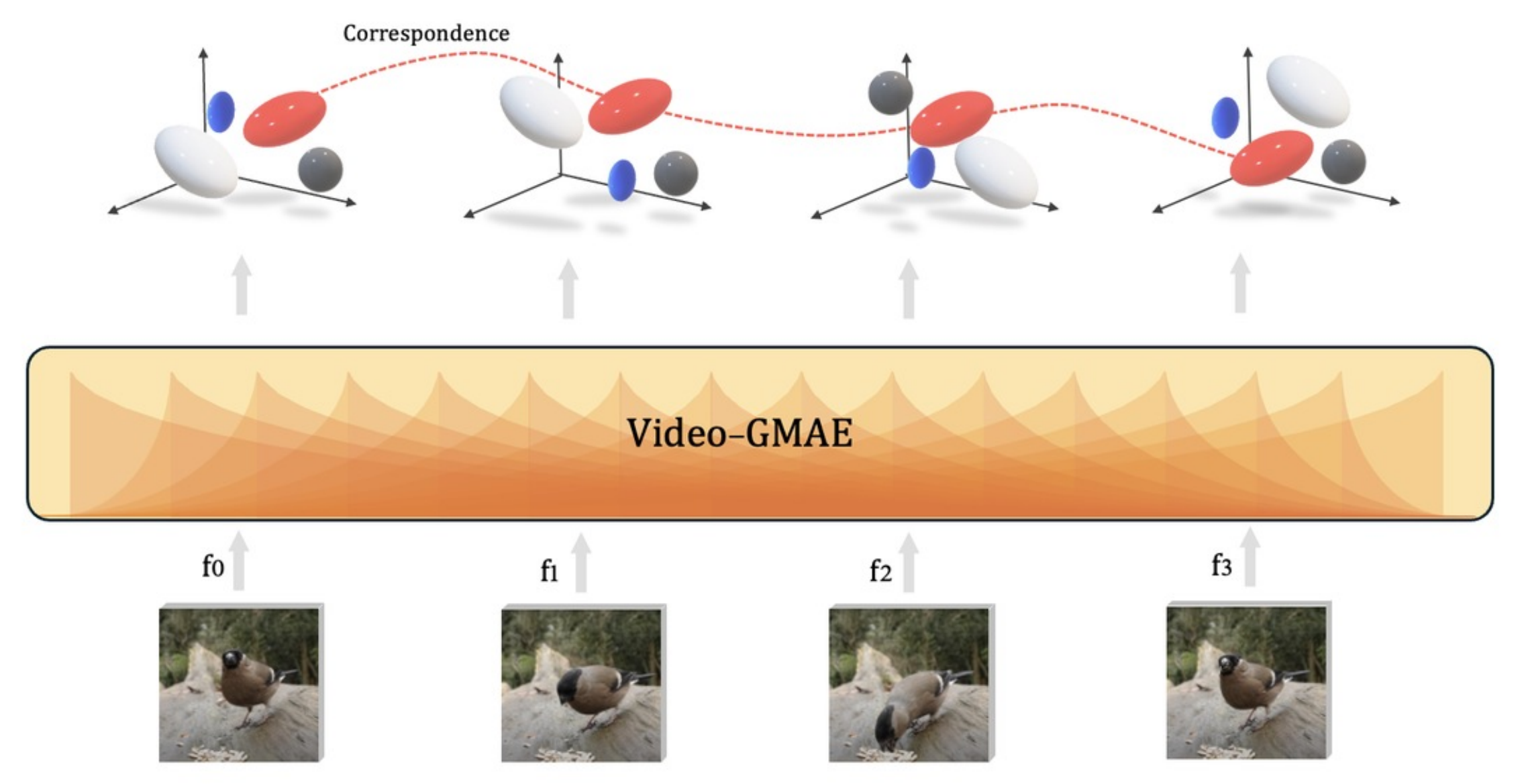}
    \caption{\textbf{Self-supervised Video Pretraining for Correspondence:} Given a sequence of video frames, our approach \approachName{x} predicts Gaussian primitives for each frame to reconstruct the whole video. In addition to this, we also enforce correspondence in the Gaussian primitives by predicting the delta Gaussians for all but the first frame.}
    \label{fig:intro}
\end{figure*}

\section{Related Work}

\noindent \textbf{Self-supervised Learning:} Over the years, self-supervised pretraining has shown strong performance across domains like language, vision, and robotics. In computer vision, there are broadly two schools of thought: discriminative and reconstructive pretraining. Discriminative methods typically train a model to recognize that different augmentations of the same image should be close in feature space. Early works like \citep{wu2018unsupervised} and SimCLR~\cite{chen2020simple} demonstrated that contrastive learning over such instance discrimination tasks can yield strong visual features. Later methods like MoCo~\cite{he2020momentum} and DINO~\cite{caron2021emerging} further pushed this line of work, showing that such learned representations can transfer well to a range of downstream tasks.

Reconstructive pretraining learns to model the data distribution by trying to reconstruct an image or a video from its noisy version. One of the most successful methods for such pretraining in computer vision has been the BERT~\citep{Devlin2018}-style masked modeling of images proposed by BEiT~\citep{bao2021beit}, and MAE~\citep{he2022masked}. Compared to BERT, MAE uses asymmetric encoder-decoders, allowing it to be very efficient at training with high masking ratios. This style of reconstructive pre-training learns strong visual priors and shows impressive results on various downstream tasks such as object detection~\citep{li2022exploring}, pose estimation~\citep{xu2022vitpose}, and robot tasks~\citep{radosavovic2023real}. Extending this to videos, VideoMAE~\cite{wang2023videomae} and MAE-ST~\cite{Feichtenhofer2022} showed that, with large masking ratios, MAEs can learn very strong representations from unlabeled videos. Another line of generative video modeling uses autoregressive models to simply predict the next patch or next frame~\cite{rajasegaran2025empirical, ren2025beyond}. Recently, these models have been used as an encoder for vision-language models~\cite{tong2024cambrian, fini2024multimodal}.


\medskip
\noindent \textbf{Gaussian Splatting:}
Gaussian Splatting~\cite{Kerbl2023} is a recent differentiable rendering method that uses 3-D Gaussian primitives as the underlying representation, allowing for flexible optimization and high-quality reconstructions. This idea builds on a broader trend in differentiable rendering, which has become a popular way to connect 3-D geometry with 2-D image supervision. By making the rendering process differentiable, these methods enable gradient-based learning of 3-D structures, like meshes and point clouds, from images. For instance, \citep{liu2019soft} introduced a soft rasterizer for mesh-based rendering, while \citep{Lassner_pulsar} proposed an efficient differentiable renderer for large point clouds. NeRF~\cite{mildenhall2021nerf} and Mip-NeRF~\cite{barron2022mip} extend this idea to volumetric scene representations, using differentiable volume rendering~\cite{levoy1990efficient} to learn 3-D radiance fields from just a few multi-view images.



\medskip
\noindent \textbf{Point Tracking:} Tracking has been studied in computer vision under different scales. Traditional methods like the Kanade–Lucas–Tomasi tracker~\cite{tomasi1991detection} have been widely used for this purpose, leveraging local image gradients to track features over time. Recent deep-learning driven advancements have introduced more robust and flexible approaches. For instance, RAFT \cite{teed2020raft} extracts per-pixel features per-frame and uses correlation across frames to compute point tracks. The Tracking Any Point (TAP) ~\cite{doersch2022tap} paradigm focuses on tracking arbitrary points on deformable surfaces, accommodating complex motions and occlusions. Models like TAPIR~\cite{doerry2023tapir} enhance this capability by employing per-frame initialization and temporal refinement strategies, enabling accurate tracking of points across diverse scenarios.  Both the above methods are trained using supervised learning with synthetic data. Another line of work focuses on developing SSL algorithms for point tracking. CRW \cite{jabri2020crw} models the evolution of patches over time as a random walk parameterized by a learnable matrix, trained using cycle consistency. GMRW \cite{shrivastava2024gmrw} further extends it to pixel-level tracking. DIFT \cite{luo2021dift} shows that nearest neighbour on patch-level features extracted from pretrained diffusion models provides not just temporal, but also semantic correspondence across frames. 

Benchmark datasets like TAP-Vid~\cite{doersch2022tap} and TAPVid-3-D~\cite{koppula2024tapvid} have been developed to evaluate and compare the performance of various point tracking methods, providing standardized metrics and diverse testing scenarios. Overall, the evolution of point tracking techniques, from classical methods to modern deep learning-based approaches, reflects the ongoing efforts to achieve more accurate, robust, and versatile tracking capabilities in computer vision applications.

\section{Preliminaries}

\subsection{Self-supervised Masked Autoencoders}

Masked autoencoders learn data representations by randomly masking parts of the input and training the model to predict the missing content. In language, BERT \cite{Devlin2018} follows this approach by using a transformer \cite{Vaswani2023} to predict masked text tokens. For images, methods like MAE \cite{He2021} and BEiT \cite{Bao2022} train the model to reconstruct masked image patches. In videos, approaches such as VideoMAE \cite{Tong2022} and MAE-ST \cite{Feichtenhofer2022} extend this idea by masking spatio-temporal patches across frames. MAE-ST uses a Vision Transformer (ViT)~\cite{Dosovitskiy2021} encoder to process the visible patches, and a lightweight ViT decoder that ingests visible and masked tokens to reconstruct the missing video content.

\subsection{3D Gaussian Splatting}
3-D Gaussian Splatting originally introduced for optimization-based single-scene 3-D reconstruction~\cite{Kerbl2023} was later extended to image-level representation learning in~\citep{Rajasegaran24}. Each primitive is defined by a 3D center position \( p \in \mathbb{R}^3 \), a covariance matrix \( \Sigma \in \mathbb{R}^{3 \times 3} \), a color \( r \in \mathbb{R}^3 \), and an opacity value \( o \in \mathbb{R} \), collectively encoding the geometry and appearance of the scene. During rendering, the Gaussians are transformed into the camera frame and projected onto the image plane using volumetric splatting. This rendering pipeline is fully differentiable, allowing gradients to flow back to the Gaussian parameters from the rendered output. Following standard practice, the covariance matrix is factorized as \( \Sigma = R S S^T R^T \), where \( S = \mathrm{diag}(s) \in \mathbb{R}^{3 \times 3} \) is a diagonal scaling matrix parameterized by \( s \in \mathbb{R}^3 \), and \( R \in \mathbb{R}^{3 \times 3} \) is a rotation matrix represented via a quaternion \( \phi \in \mathbb{R}^4 \). As a result, each Gaussian is fully described by a 14-dimensional vector \( g = \{\mu, s, \phi, r, o\} \in \mathbb{R}^{14} \).
\vspace{0.5em}
\section{Self-Supervised Pretraining}

\begin{figure*}[!tb]
  \centering
    \includegraphics[width=0.7\linewidth]{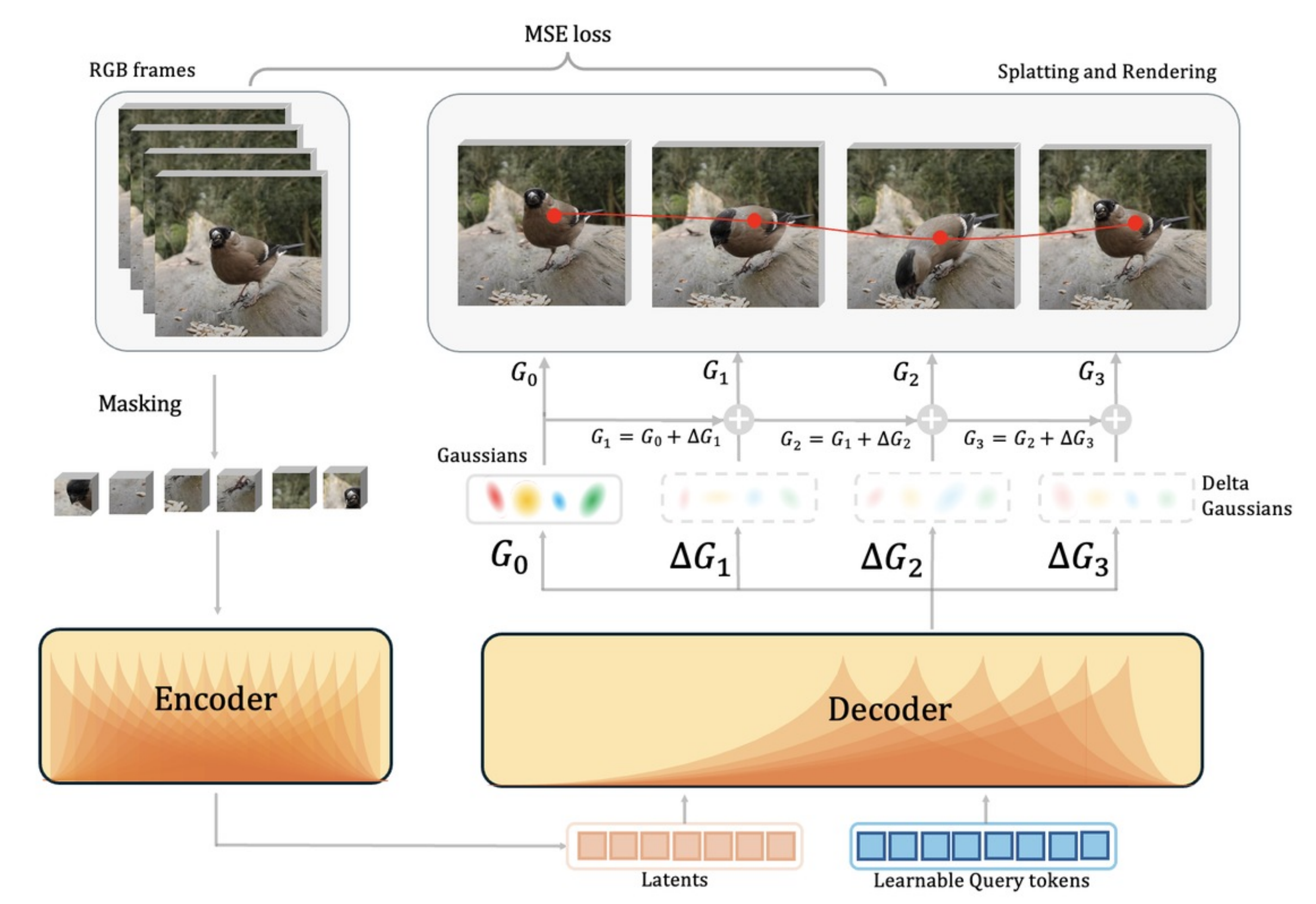}
   \caption{\textbf{Video Masked Auto encoding via Gaussian Splatting:} The ViT Encoder processes masked input frames to produce latent embeddings. The ViT Decoder then predicts explicit Gaussian parameters for frame $f_1$ based on query tokens, including color, opacity, center, scale, and orientation, and the Gaussian deltas for the rest of the frames. The explicit Gaussians for frame $f_1$ to $f_t$ are calculated and rendered via differentiable volume splitting to reconstruct all the frames. We pre-train our models fully end-to-end with self-supervision.}
   \vspace{-0.2cm}
   \label{fig:methods}
\end{figure*}

Our self-supervised pretraining method is summarized in Fig~\ref{fig:methods}. We feed $k\!=\!16$ RGB frames into a ViT encoder as video patches with a masking ratio of 95\% as in ~\citep{Feichtenhofer2022}. The encoder computes latents, which are passed into a decoder concatenated with query tokens. The decoder generates $k \times n$ Gaussian primitives for rendering. The first $n$ of these are treated as independent Gaussian primitives in free space for the first frame. The rest are \textit{residual} Gaussians deltas, which are integrated to compute Gaussians for subsequent frames. Predicting residuals for the first set of Gaussians enforces correspondence in the Gaussian primitives over time.

We start with an image-based encoder trained to predict 3-D Gaussians for an image~\citep{Rajasegaran24}. We extend this model to be a video encoder by adding a separate, learnable positional embedding for the time axis. We train this video model on $k$ frames, with $16\times16$ patches per frame. We do patchification at the frame level to utilize the image model--~GMAE~\citep{Rajasegaran24}.

The decoder takes in encoder latents and $k \times n$ learnable query tokens, $k\!\!=\!\!16$ frames at pretraining, and $n\!\!=\!\!256$ Gaussians per frame. Define $G_0 = \{g_1, g_2, ... g_n\}$ as the set of Gaussians for the first frame. For subsequent frames, the decoder only predicts delta Gaussians, $\Delta G_1, ..., \Delta G_k$. We find that predicting $\Delta G = \big\{ \{\Delta \mu_1, \Delta r_1\}, \dots,\{\Delta \mu_n, \Delta r_n\}\big\}$ is sufficient. During pretraining, we integrate the delta Gaussians per frame $G_t = \Delta G_{t} + G_{t-1}$ to get the $k$ sets of Gaussians primitives that we render using Gaussian splatting, \cite{Kerbl2023} and train the model end-to-end with reconstruction loss. This formulation allows us to train the model with correspondence under self-supervised training.

Our pretraining dataset combines the training sets of Kinetics\cite{kay2017kinetics}, Kubric \cite{greff2022kubric} and Davis \cite{perazzi2016davis}. \textit{We only use the unlabeled videos from these datasets.} More details about pre-training methodology can be found in the supplementary.

\section{Zero-shot Point Tracking}

\begin{figure}[h]
\centering
\includegraphics[width=0.5\linewidth]{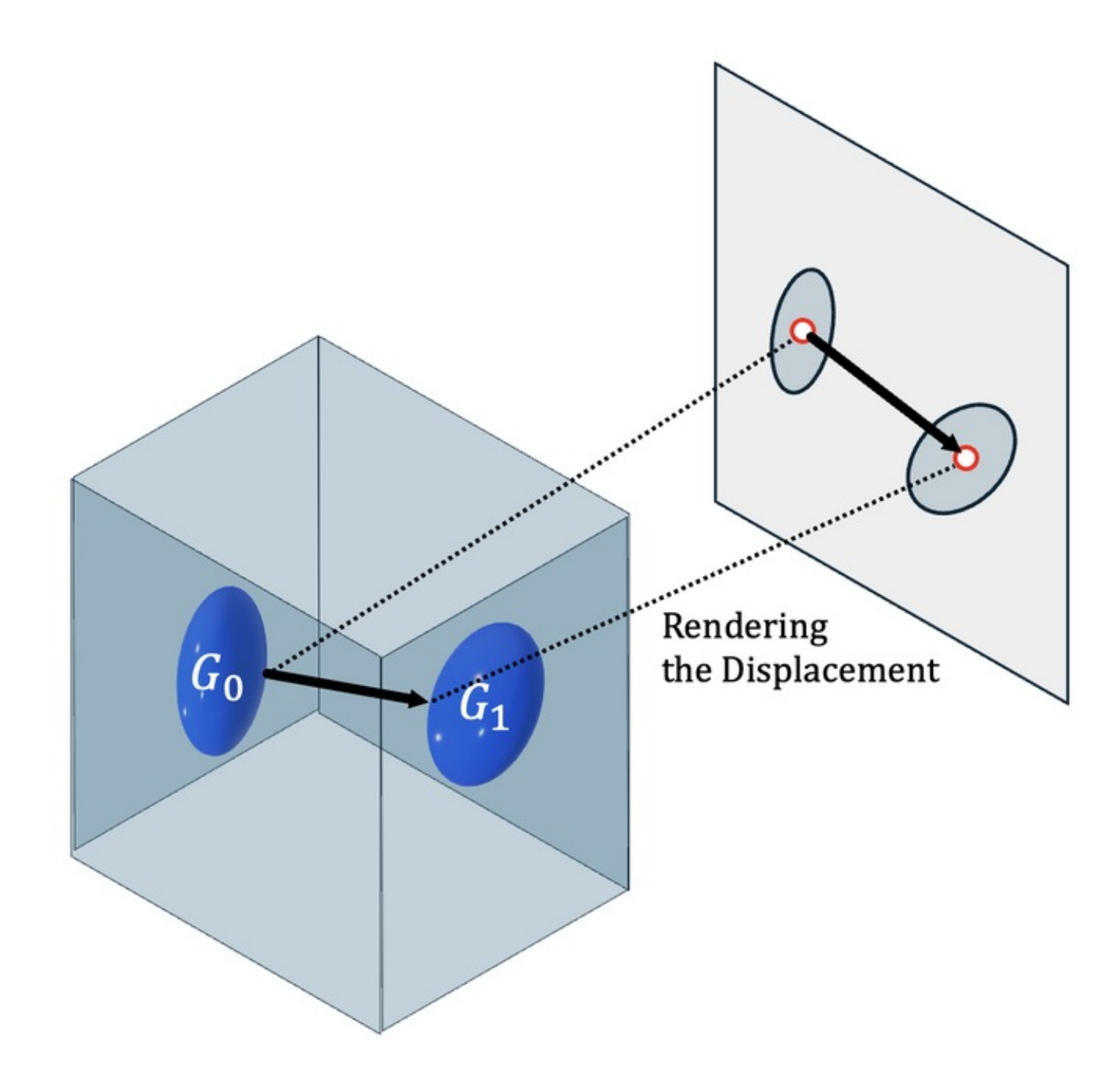}
\caption{\textbf{Zero-shot Point Tracking:} The 3-D centers of the predicted Gaussian primitives are projected and the subsequent 2-D displacement vector is rendered.}
\label{fig:zero_shot}
\end{figure}

In this section, we present an algorithm to extract point tracks from the collection of Gaussian primitive trajectories predicted by \approachName{}. At a high level, we splat the Gaussian primitives on the image plane with the RGB values of each Gaussian replaced with the projected 2-D $\Delta \mu_i$. This renders the 2-D Gaussian mean deltas into a flow field. Following the flow for any point tracks it to the next frame.

Given an initial set of Gaussians $G_0 = \{g_1, g_2, \dots, g_n\}$ for frame $t=0$, each primitive $g_i = \{\mu_i^{(0)}, s_i, \phi_i, r_i^{(0)}, o_i\}$ contains a 3-D position $\mu_i\in \mathbb{R}^3$. For subsequent frames, the model predicts residual updates $\Delta G_t = \{\Delta \mu_i^{(t)}, \Delta r_i^{(t)}\}_{i=1}^n$ to the Gaussian means. We apply these deltas recursively to get the means of the Gaussian primitives as:
\begin{equation}
\mu_i^{(t+1)} = \mu_i^{(t)} + \Delta \mu_i^{(t)}.
\end{equation}
To extract motion in the image plane, we project the means of each Gaussian into pixel coordinates using the same camera intrinsics $K \in \mathbb{R}^{3 \times 3}$ and extrinsics $[R \mid t] \in \mathrm{SE}(3)$, that we use to render during the training phase. Let $\Pi(\cdot)$ denote the perspective projection:
\begin{equation}
x_i^{(t)} = \Pi\left( K [R \mid t], \mu_i^{(t)} \right).
\end{equation}
We also use the projected Gaussian locations to define a per-Gaussian image-plane offset:
\begin{equation}
\Delta x_i^{(t+1)} = x_i^{(t+1)} - x_i^{(t)}.
\end{equation}

We calculate the instantaneous 2-D displacement vector $\Delta x_i^{(t)}$ carried by each Gaussian, encode them as pseudo-RGB values $c_i^{(t)} = (\Delta x^{(t)}_{i,x}, \Delta x_{i,y}^{(t)}, 0)$, and splat them onto the image plane using standard volumetric alpha compositing. The resulting dense flow field $F^{(t)} \in \mathbb{R}^{H \times W \times 2}$ at each pixel $u \in \mathbb{R}^2$ is computed by an opacity-weighted average of all displacements:
\begin{equation}
F^{(t)}(u) = \sum_{i=1}^n \alpha_i^{(t)}(u)\cdot d_i^{(t)},
\end{equation}
where $\alpha_i^{(t)}(u) \in [0,1]$ is the rasterized visibility of Gaussian $i$ in the pixel $u$, calculated using differentiable Gaussian splatting \cite{Kerbl2023}. We extend this notation for an arbitrary point $p$ using bilinear interpolation, where $\alpha_i^{(t)}(p) = \mathrm{bilinear}\big(\alpha_i^{(t)}, p\big)$ and $F^{(t)}(p) = \mathrm{bilinear}\big(F^{(t)}, p\big)$.

To improve robustness around occlusions, we additionally leverage the renderer’s soft per-pixel assignments. First, we fix a per-point anchor set $\mathcal{S}=\mathrm{Top_k}\!\big(\big\{\alpha_j^{(t)}(u)\big\}_{j=1}^n\big)$ chosen once at $t=0$. Then, we define the per-frame anchor mass:
\begin{equation}
\omega^{(t)} \;=\; \textstyle\sum_{i\in\mathcal{S}} \alpha_i^{(t)}\big(p^{(t)}\big), 
\end{equation}
and re-normalized mixture weights over this fixed set:
\begin{equation}    
\tilde{\pi}_i^{(t)} \;=\; \dfrac{\alpha_i^{(t)}\big(p^{(t)}\big)}{\sum_{j\in\mathcal{S}} \alpha_j^{(t)}\big(p^{(t)}\big) + \varepsilon},\; i\!\in\!\mathcal{S}.
\end{equation}
The point is visible if $\omega^{(t)} \ge \tau_{\mathrm{vis}}$ and occluded otherwise.

\newcommand{\A}[1]{a^{(#1)}}           
\newcommand{\Svis}[1]{s^{(#1)}}        
\newcommand{\Soccl}[1]{\bar{s}^{(#1)}} 

We write $\A{t} := p^{(t)} + F^{(t)}\!\big(p^{(t)}\big)$ for $p^{(t)}$ advected by $F^{(t)}$ and $\Svis{t{+}1} := \sum_{i\in\mathcal{S}} \tilde{\pi}_i^{(t)}\!\Big(x_i^{(t)} + \Delta x_i^{(t+1)}\Big)$ for the weight-averaged anchor Gaussian primitives plus their displacements. Then, the per-frame update for $p^{(t)}$ is:
\begin{equation}
p^{(t+1)}=
\begin{cases}
(1{-}\beta)\A{t} + \beta\,\Svis{t{+}1} & \omega^{(t)} \ge \tau_{\mathrm{vis}}\\[4pt]
\Svis{t{+}1} & \omega^{(t)} < \tau_{\mathrm{vis}}.
\end{cases}
\end{equation}
Here $\beta\!\in\![0,1]$ blends pure flow-based advection with the top-k Gaussian-mixture proposal. Setting $k = n$ and $\tau_{\text{vis}}=0$ recovers a pure flow-based update that ignores occlusions. If the updated location $p^{(t+1)}$ lies outside the image dimensions, we mark the point as occluded. The hyperparameters $k=8, \tau_{\text{vis}}=0.5,$ and $\beta=0.3$ are determined based on performance on the Kubric train dataset.

\section{Supervised Finetuning for Point Tracking}

\begin{figure*}[!tb]
  \centering
    \includegraphics[width=1\linewidth]{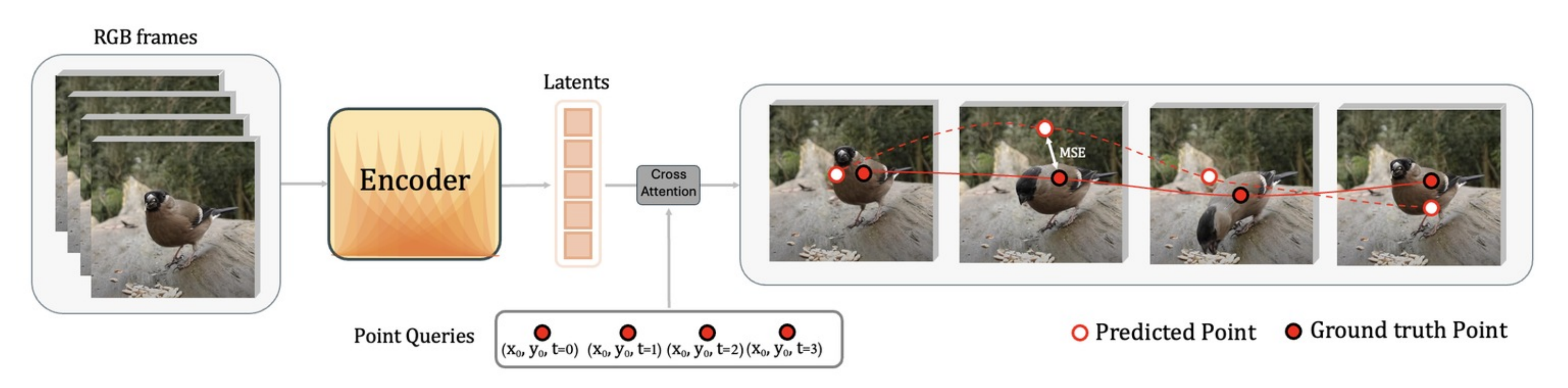}
   \caption{\textbf{Finetuning for Point Tracking:} We use our pretrained encoder without masking, and query the latents to predict the best point tracks. We finetune this model, using the annotated Kubric~\cite{greff2022kubric} dataset. The Fourier embeddings of the initial queries are calculated, and they cross-attend to the encoder latents in the fine-tuned cross-attention readout layer.}
   \label{fig:finetune-methods}
\end{figure*}

We evaluate the encoder's learned representation on point tracking to assess the improved correspondence-based features. For point-tracking evaluation, we adopt a cross–attention–based readout architecture and evaluate it on the TAP-Vid protocol, which includes three datasets: TAP-Vid-Kinetics\cite{doersch2022tap}, TAP-Vid-DAVIS\cite{doersch2022tap}, and Kubric\cite{greff2022kubric}. Each evaluation run uses a single query per point track. We follow the evaluation protocol described in \citep{shrivastava2024gmrw}. We either freeze (\approachName{}-frozen) or fine-tune (\approachName{}-finetune) the encoder, and the readout network consumes the spatio-temporal encoder features to predict tracked points across frames, as shown in Fig~\ref{fig:finetune-methods}.

Following the design in \citep{Carreira24}, we first apply layer normalization to the encoder features, add learnable temporal embeddings, and use 64-dimensional Fourier-based positional queries per frame, projected to the token feature dimension. These queries cross-attend to the processed encoder features using a 16-headed attention mechanism. This is followed by a residual MLP with hidden size four times the token feature dimension and GeLU \cite{hendrycks2023gelu} activation, and finally a linear layer mapping with sigmoid activation to a 3-D vector containing the tracked 2-D point and its occlusion.

For frozen evaluation, we train only the readout layers. We train each model on one A100, with batch size of 8, and 16 tracks per video for 50k steps, using the AdamW optimizer (weight decay 5e-2) and learning rate sweep over \{5e-5, 1e-4, 3e-4\}, selecting the best-performing configuration. In the finetuned setting, the same setup is used, except both the encoder and the readout are trained end-to-end.

\section{Experiments}

\noindent\textbf{Evaluation:} For point-tracking evaluation, we report three metrics that collectively measure accuracy, robustness, and occlusion handling. Average Jaccard Score (AJ) computes the intersection-over-union between predicted and ground-truth visible regions, reflecting alignment quality over time. Average Points within Threshold ($\delta_{\text{avg}}^x$) reports the percentage of predictions falling within a small pixel radius of the ground truth, offering an interpretable precision metric. Occlusion Accuracy (OA) measures the frequency with which the model accurately predicts whether a point is visible or occluded in each frame. We conduct a series of experiments to evaluate the effectiveness and design choices of our proposed model. Unless otherwise specified, all experiments are performed with a frozen encoder, focusing exclusively on optimizing the lightweight decoder. 

\noindent\textbf{Implementation Details:} We pretrained the models on 64 V100s for 90 epochs with a batch size of 128, a learning rate of 1e-3, using the AdamW \cite{loshchilov2019adamw} optimizer (weight decay 5e-2). We used 2000 warm-up steps and cosine decay for the learning rate, and gradient clipping with a value of 2.0.

\subsection{Zero-shot Tracking Results}
\begin{figure}[!tb]
    \centering
    \includegraphics[width=0.99\linewidth]{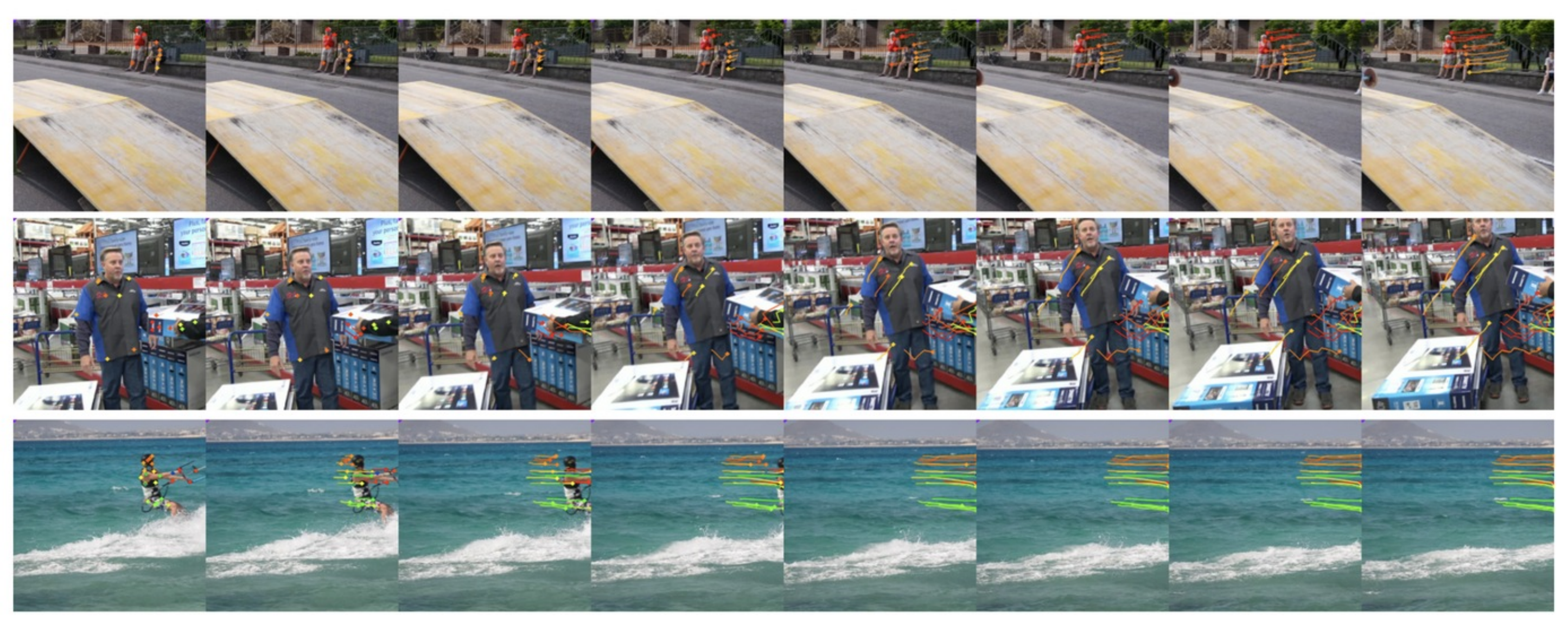}
    \caption{\textbf{Zero shot Qualitative Results:} We show qualitative results (16 frames, we only show every other frame here) from our \approachName{X}-zeroshot method. Without any track labels, the model can track objects with camera motion and pose changes, and shows robust tracking over long videos. }
    \label{fig:results_zeroshot}
\end{figure}

We compute AJ, $\delta_{\text{avg}}^x$, and OA for zero-shot tracking on all three datasets following~\cite{doersch2022tap}. Table \ref{tab:tracking_comparison} shows the zero-shot tracking metrics. Fig~\ref{fig:results_zeroshot} shows qualitative examples of the zero-shot tracking. Our zero-shot tracking results on Kubric and TAP-Vid Davis are comparable to GMRW-C \cite{shrivastava2024gmrw} numbers and outperform all self-supervised methods on TAP-Vid Kinetics. The zero-shot results show the benefit of our self-supervised approach, which is comparable to other self-supervised approaches while being scalable.

\begin{figure}[!tb]
    \centering
    \includegraphics[width=0.99\linewidth]{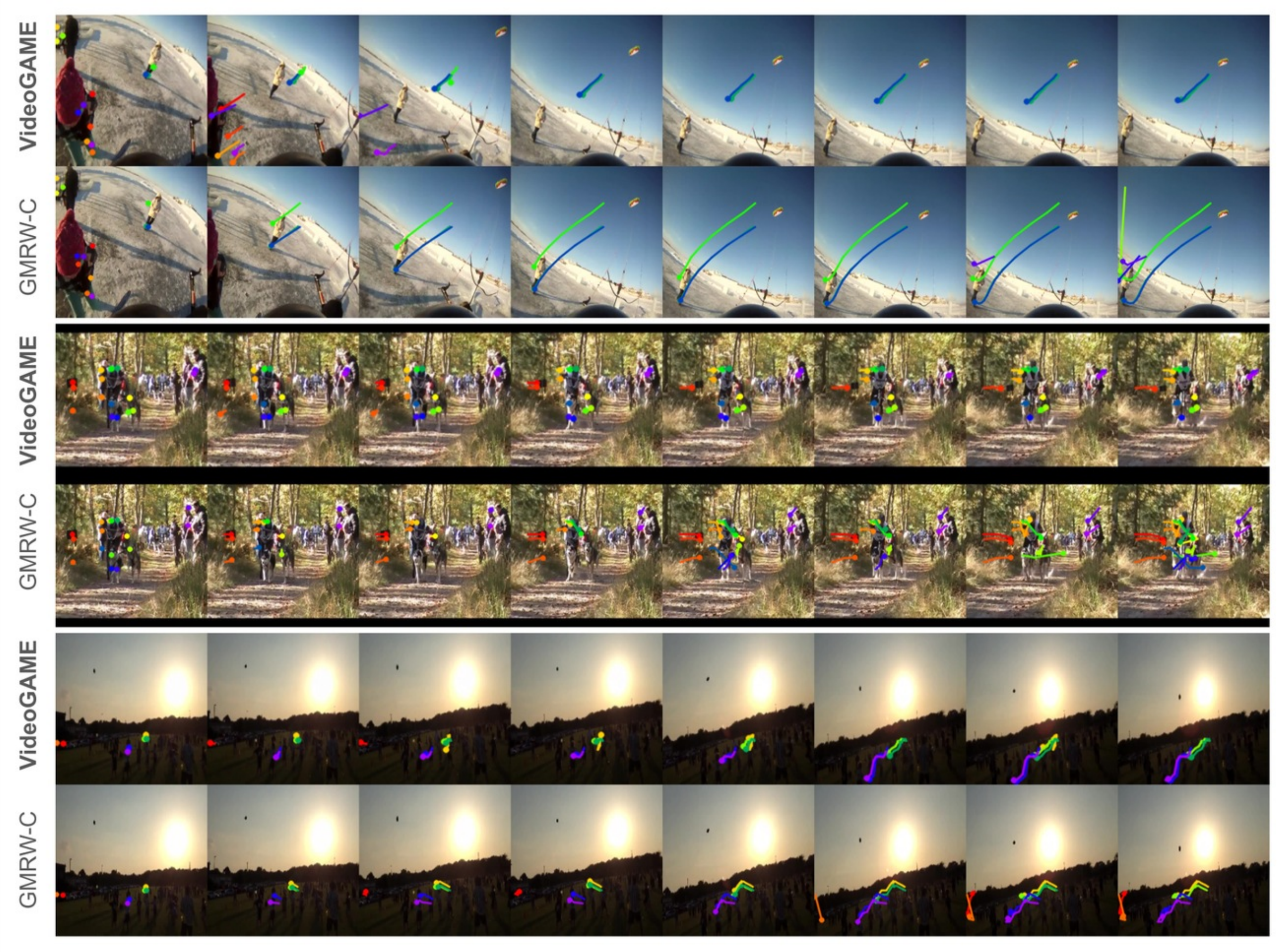}
    \caption{\textbf{Zero shot Comparison with GMRW-C:} We show qualitative comparisons between \approachName{}-zeroshot and GMRW-C~\cite{shrivastava2024gmrw} on challenging sequences. \approachName{} produces more temporally stable tracks, with fewer flickers and smoother motion, and tends to keep points visible until they leave the frame, while GMRW-C often occludes points 1--3 frames early. In the first example, where small, visually detailed regions move large distances across the frame, GMRW-C tracks them more accurately than our model, highlighting a limitation of our approach.}
    \label{fig:comparison_zeroshot}
\end{figure}

To better understand the behavior of our zero-shot tracker relative to the strongest self-supervised baseline, we conduct a qualitative comparison between \approachName{}-zeroshot and GMRW-C~\cite{shrivastava2024gmrw} on challenging sequences. Fig~\ref{fig:comparison_zeroshot} shows hand-picked examples from TAP-Vid Kinetics that highlight both strengths and limitations of our approach. Overall, we observe that \approachName{}-zeroshot produces more temporally stable tracks; once a point is visible, our method tends to persist it with minimal flickering, while GMRW-C often exhibits frame-to-frame identity switches or short-lived disappearances. \approachName{}-zeroshot is also more conservative around occlusions; it predicts tracks that are typically maintained as visible until they fully leave the frame (or sometimes one to two frames thereafter), whereas GMRW-C frequently marks points as occluded 1--3 frames early and sometimes re-activates them spuriously, which degrades OA. Also, we observe that GMRW-C has a tendency to jitter points around their true locations, while \approachName{} yields smoother, more conservative motion estimates. We also find that GMRW-C can be advantageous in certain regimes. When small, visually detailed regions in the video undergo large displacements, GMRW-C maintains their trajectories better than our model.  Modeling each scene with only 256 Gaussian primitives constrains the spatial resolution at which fine details can be represented and tracked. This behavior is consistent with a current limitation of our Gaussian representation.  These qualitative differences complement the quantitative metrics and suggest that our Gaussian-based zero-shot tracker is particularly well-suited for stable, occlusion-aware tracking, while future work could focus on improving fine-detail fidelity under large motions. More comparisons and tracking samples can be found in the supplementary.

\begin{figure}[!tb]
    \centering
    \includegraphics[width=0.99\linewidth]{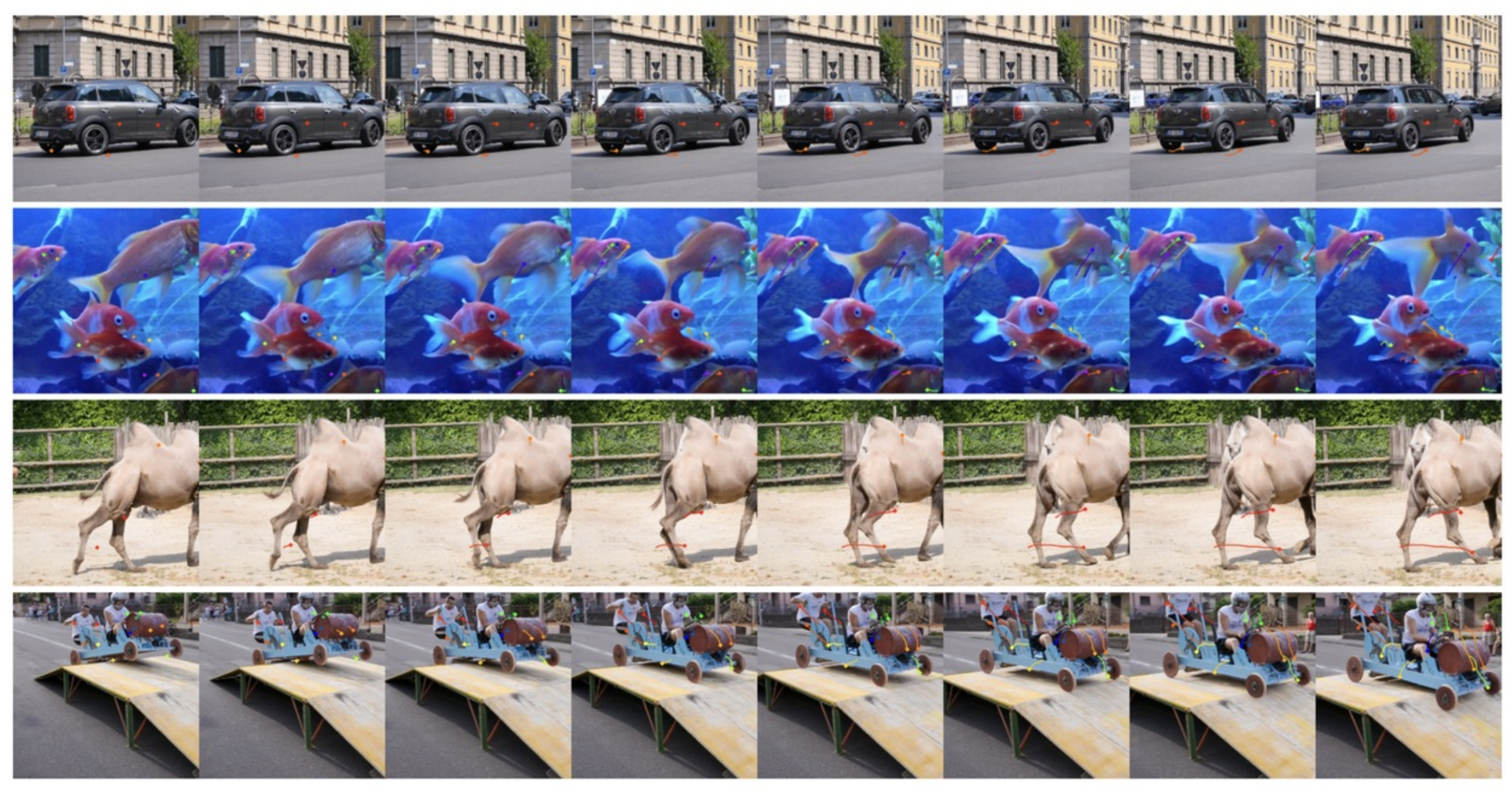}
    \caption{\textbf{Qualitative Results from Finetuned Model:} We show qualitative results (16 frames, we only show every other frame here) from our \approachName{X}-base encoder finetuned model, after finetuning on Kubric datasets~\cite{greff2022kubric}. Our model was able to achieve state-of-the-art performance on point tracking, and was able to track points over long range with high precision.}
    \label{fig:results_finetune}
\end{figure}

\subsection{Comparison with Video Pretraining Methods}
We compare \approachName{} with existing state-of-the-art self-supervised pretraining methods on the task of tracking. Specifically, we compare against VideoMAE \cite{Tong2022} and MAE-ST \cite{Feichtenhofer2022}. Following the encoder and decoder configurations of the MAE Base and MAE Large architectures, we train two versions of our model: \approachName{}-base and \approachName{}-large, respectively. These are compared against VideoMAE and MAE-ST using the same frozen-encoder training setup. Table \ref{tab:model_comparison_frozen} contains our results. \approachName{} outperforms the other pretraining baselines on all datasets.

\begin{table*}[!tb]
  \centering
  \small
  \setlength{\tabcolsep}{8pt}
  \begin{tabular}{l|ccc|ccc|ccc}
    \toprule
    \multirow{2}{*}{Model} &
    \multicolumn{3}{c|}{\textbf{Kinetics}} &
    \multicolumn{3}{c|}{\textbf{DAVIS}} &
    \multicolumn{3}{c}{\textbf{Kubric}} \\
    \cmidrule(lr){2-4}\cmidrule(lr){5-7}\cmidrule(lr){8-10}
      &
      AJ$\!\uparrow$ & $\!\delta_{\text{avg}}^{x}\!\uparrow$ & OA$\!\uparrow$ &
      AJ$\!\uparrow$ & $\!\delta_{\text{avg}}^{x}\!\uparrow$ & OA$\!\uparrow$ &
      AJ$\!\uparrow$ & $\!\delta_{\text{avg}}^{x}\!\uparrow$ & OA$\!\uparrow$ \\
    \midrule
    
    MAE-ST\cite{Feichtenhofer2022}  & 42.3 & 49.8 & 95.4 & 28.3 & 37.2 & 90.1 & 41.5 & 51.6 & 95.9 \\
    VideoMAE\cite{Tong2022}  & 46.9 & 54.3 & 94.9 & 31.8 & 39.8 & 90.4 &  44.8 & 55.2 & 95.8 \\
    \approachName{}     & \textbf{65.1} & \textbf{72.0} & \textbf{97.4} & \textbf{46.7} & \textbf{55.7} & \textbf{91.8} & \textbf{62.4} & \textbf{71.9} & \textbf{96.6} \\
    \bottomrule
  \end{tabular}
  \caption{Cross-dataset comparison of pre-trained backbones (all are with a frozen encoder) comparing \approachName{} to VideoMAE and MAE-ST. Even though we use the same masked auto-encoding, our correspondence-aware decoder forces the model to learn better representations for point-tracking. We compare using a stride of five frames, the same as in Table \ref{tab:tracking_comparison}.}
  \label{tab:model_comparison_frozen}
\end{table*}

\begin{table*}[!tb]
  \centering
  \small
  \setlength{\tabcolsep}{4pt}
  \begin{tabular}{l | ccc | ccc | ccc}
    \toprule
    \multirow{2}{*}{Method} &
    \multicolumn{3}{c}{\textbf{Kubric}} &
    \multicolumn{3}{c}{\textbf{DAVIS}} &
    \multicolumn{3}{c}{\textbf{Kinetics}} \\
    \cmidrule(lr){2-4}\cmidrule(lr){5-7}\cmidrule(lr){8-10}
      & AJ $\uparrow$ & $\langle\delta^{x}_{\text{avg}}\rangle\uparrow$ & OA $\uparrow$
        & AJ $\uparrow$ & $\langle\delta^{x}_{\text{avg}}\rangle\uparrow$ & OA $\uparrow$
        & AJ $\uparrow$ & $\langle\delta^{x}_{\text{avg}}\rangle\uparrow$ & OA $\uparrow$ \\
    \midrule
    \multicolumn{10}{l}{\textit{Supervised}}\\
    Kubric-VFS-Like~\cite{kuhn2023vfs}        & 51.9 & 69.8 & 84.6 & 33.1 & 48.5 & 79.4 & 40.5 & 59.0 & 80.0 \\
    COTR~\cite{shi2021cotr}                   & 40.1 & 60.7 & 78.6 & 35.4 & 51.3 & 80.2 & 19.0 & 38.8 & 57.4 \\
    TAP-Net~\cite{dwibedi2019tapnet}          & 65.4 & 77.7 & 93.0 & 38.4 & 53.1 & 82.3 & 46.6 & 60.9 & 85.0 \\
    PIPs~\cite{harley2022pips}                & 59.1 & 74.8 & 88.6 & 42.0 & 59.4 & 82.1 & 35.3 & 54.8 & 77.4 \\
    TAPIR~\cite{doerry2023tapir}              & -- & -- & -- & 56.2 & 70.0 & 86.5 & 49.6 & 64.2 & 85.0 \\
    CoTracker~\cite{gouiaa2023cotracker}      & -- & -- & -- & 61.8 & 76.1 & 88.3 & 49.6 & 64.3 & 83.3 \\
    CoTracker3~\cite{karaev2024cotracker3}    & -- & -- & -- & \textbf{63.8} & \textbf{76.3} & 90.2 & 55.8 & 68.5 & 88.3 \\
    LocoTrack~\cite{cho2024}                  & -- & -- & -- & 62.9 & 75.3 & 87.2 & 52.9 & 66.8 & 85.3 \\
    BootsTAPIR~\cite{doersch2024bootstapir}   & -- & -- & -- & 61.4 & 73.6 & 88.7 & 54.6 & 68.4 & 86.5 \\
    PIPs++~\cite{zheng2023pipspp}             & --   & -- & --   & --   & 73.7 & --   & --   & 63.5 & --   \\
    \midrule
    \multicolumn{10}{l}{\textit{Self-supervised}}\\
    CRW-C~\cite{jabri2020crw}                 & 31.4 & 48.1 & 76.3 &  7.7 & 13.5 & 72.9 & 20.2 & 33.6 & 70.6 \\
    CRW-D~\cite{jabri2020crw}                 & 35.8 & 52.4 & 80.9 & 23.6 & 38.0 & 77.2 & 21.9 & 36.8 & 70.4 \\
    DIFT-C~\cite{luo2021dift}                 & 28.3 & 45.2 & 69.0 & 18.1 & 33.0 & 68.8 & 19.8 & 33.7 & 68.7 \\
    DIFT-D~\cite{luo2021dift}                 & 41.6 & 59.8 & 83.9 & 29.7 & 48.2 & 77.2 & 19.5 & 34.4 & 70.1 \\
    Flow-Walk-C~\cite{park2023flowwalk}       & 49.4 & 66.7 & 82.7 & 35.2 & 51.4 & 80.6 & 40.9 & 55.5 & 84.5 \\
    Flow-Walk-D~\cite{park2023flowwalk}       & 51.1 & 68.1 & 80.3 & 24.4 & 40.9 & 76.5 & 46.9 & 65.9 & 81.8 \\
    ARFlow-C~\cite{liu2020arflow}             & 52.3 & 68.1 & 81.4 & 35.0 & 51.8 & 79.7 & 27.3 & 44.3 & 79.5 \\
    GMRW-C~\cite{shrivastava2024gmrw}         & 54.2 & \underline{72.4} & 82.6 & \underline{41.8} & \underline{60.9} & 78.3 & 31.9 & 52.3 & 72.9 \\
    GMRW-D~\cite{shrivastava2024gmrw}         & 51.4 & 71.7 & 83.9 & 30.3 & 49.4 & 77.3 & 36.3 & 59.2 & 71.0 \\
    \midrule
    \approachName{X} zeroshot & \underline{54.3} & 67.0 & \underline{91.9} & 41.3 & 55.7 & \underline{85.2} & \underline{60.1} & \underline{69.1} & \underline{90.7} \\
    \approachName{X} base frozen & 60.8 & 70.7 & 96.6 & 45.2 & 55.2 & 91.4 & 61.7 & 68.9 & 97.1 \\
    \approachName{X} base finetune & 73.6 & 82.3 & 97.5 & 55.7 & 66.1 & 92.1 & 75.0 & 81.7 & 97.7 \\
    \approachName{X} large frozen & 62.4 & 71.9 & 96.6 & 46.7 & 55.8 & 91.8 & 65.1 & 72.0 & 97.4 \\
    \approachName{X} large finetune & \textbf{74.0} & \textbf{82.4} & \textbf{97.6} & 57.9 & 67.7 & \textbf{93.5} & \textbf{75.1} & \textbf{81.6} & \textbf{97.9} \\
    \bottomrule
  \end{tabular}
  \caption{Performance comparison on three video datasets. \approachName{x} shows strong performance across multiple datasets and multiple metrics. To compare with all the models, we use a stride of five for evaluations. We bold the best supervised performance metrics and underline the best self-supervised performance metrics.}
  \vspace{-0.2cm}
  \label{tab:tracking_comparison}
\end{table*}

\subsection{Comparison with Tracking Baselines}
We compare \approachName{} with state-of-the-art self-supervised methods like DIFT \cite{luo2021dift} and GMRW \cite{shrivastava2024gmrw} and supervised tracking methods like LocoTrack \cite{cho2024} and CoTracker3 \cite{karaev2024cotracker3}. We show quantitative results in Table \ref{tab:tracking_comparison}, and Figure \ref{fig:results_finetune} shows some qualitative results of the fine-tuned point tracking. To compare with other models, we use a stride of five for evaluations~\cite{doersch2022tap}. We train two models, \approachName{}-base and \approachName{}-large. For each model, we train with a frozen encoder and a fully fine-tuned encoder. \approachName{} finetune outperforms the baselines at all model scales on Kubric and TAP-Vid Kinetics benchmarks. We find that the model falls behind CoTracker \cite{gouiaa2023cotracker}, CoTracker3 \cite{karaev2024cotracker3}, LocoTrack \cite{cho2024}, and BootsTAPIR \cite{doersch2024bootstapir} for the TAP-Vid Davis benchmark. Among frozen encoder evaluations, we find that the large models outperform the base models. We see a similar scaling trend with the fine-tuned models, which outperform the frozen encoder models. To have a fair comparison, in Table \ref{tab:tracking_comparison}, we compare our \approachName{}-zeroshot against self-supervised approaches and \approachName{}-frozen and \approachName{}-finetune against the supervised approaches.

\subsection{Frame Length Scaling}
In this experiment, we investigate how our representations evolve as we train on longer frame sequences. We perform this ablation with \approachName{}-base on frame lengths \{ 2, 4, 8, 16, 24 \}. We evaluate these models on the strided TAPVid \cite{doersch2022tap} evaluation protocol, once with the stride equal to the number of frames the model was trained on, and once with a stride of two for a faithful comparison across the different ablations. We evaluate these using AJ on TAP-Vid Davis, and show results in Table~\ref{tab:ablation_combined}.

We find that training on longer frame sequences and evaluating on longer stride lengths leads to decreased AJ numbers. This is likely due to longer frame lengths inducing a stronger regularization from the correspondence-aware pre-training, which limits the quality of learned representations. When comparing on a fixed stride of two, we find that information of the future only helps with tracking to a certain extent. Having four and eight frames of information seems to help the model predict at a stride of two, but increasing it to 16 and 24 seems to decrease AJ due to the stronger regularization at these frame lengths.

\subsection{Delta Gaussian Ablations}
We also explore the importance of the delta Gaussian parameters in modeling temporal evolution. In Table~\ref{tab:ablation_combined}, we ablate three variants using \approachName{}-large: (1) integrating only the mean components $\Delta \mu_{ij}$, (2) integrating only the RGB components $\Delta r_{ij}$, and (3) integrating neither of them, and having static Gaussians. This helps isolate the contribution of motion versus appearance changes in temporal modeling.  By evaluating AJ on TAP-Vid Davis, we find that integrating mean correspondence not only enables zero-shot point-tracking but also improves the learned encoder latents for fine-tuned point-tracking. 

\begin{table}[!tb]
  \centering
  \small
  \setlength{\tabcolsep}{4pt}
  \renewcommand{\arraystretch}{1.1}
  \begin{tabular}{ccc|ccc}
    \toprule
    \multicolumn{3}{c|}{Delta Ablation} &
    \multicolumn{3}{c}{\# Frames vs. Stride} \\
    $\Delta \mu_{ij}$ & $\Delta r_{ij}$ & AJ $\uparrow$ &
    \# Frames (F) & Stride = F & Stride = 2 \\
    \midrule
    \checkmark & \checkmark & 44.7 & 2  & 55.4 & 55.4 \\
    \checkmark & --         & 44.4 & 4  & 49.2 & 64.2 \\
    --         & \checkmark & 42.5 & 8  & 41.5 & 63.5 \\
    --         & --         & 39.1 & 16 & 47.8 & 57.5 \\
               &            &      & 24 & 24.7 & 52.4 \\
    \bottomrule
  \end{tabular}
  \caption{Ablations on finetuned tracking. Left: effect of using mean and RGB deltas during pretraining on Davis AJ. Right: impact of the number of frames and stride on Davis AJ for \approachName{}-base, comparing stride equal to the number of frames (F) and a fixed stride of two.}
  \label{tab:ablation_combined}
\end{table}

\vspace{-0.2cm}
\section{Limitations}
\vspace{-0.2cm}

One of the main limitations of our work is the modeling of the camera. Throughout the pre-training, we assume a static camera, but this is not an ideal assumption when pretraining on internet-style videos. Because of the assumption about the static camera, we also fail to recover any metric 3-D information in the 3-D Gaussians. Another assumption we made in this work is regarding the correspondence-based regularization. While this is a weaker assumption than the prior one, and works reasonably well on short videos, when the number of frames at pre-training becomes very large, this regularization starts to hurt the learning. Additionally, during the pretraining, we only use 256 Gaussians per frame, which significantly limits the quality of our renderings. These limitations also restrict the quality of the Gaussian representation; addressing them should yield even better zero-shot tracking performance.

\vspace{-0.2cm}
\section{Societal Impact}
\vspace{-0.2cm}

Our work introduces a self-supervised approach to learning point correspondence in videos by predicting 3D Gaussian trajectories, enabling robust zero-shot tracking without relying on human annotations. This has positive societal implications by reducing the dependency on costly, labor-intensive labeled data, which can democratize access to high-quality video understanding models in domains such as robotics, assistive technologies, and environmental monitoring. However, as with any tracking technology, it also presents potential risks related to privacy and surveillance, especially if misused in contexts lacking consent or oversight. While our model does not include person identification and assumes static cameras during training, we recognize the broader ethical implications and emphasize the importance of responsible deployment, transparency, and alignment with legal and ethical norms in real-world applications.

\vspace{-0.2cm}
\section{Conclusion}
\vspace{-0.2cm}

In this paper, we introduced \approachName{X} for self-supervised learning from videos with built-in correspondence. By predicting the small changes in the Gaussian primitives over time, we enforced correspondence in videos. This also shows an alternative to patch-based video pretraining approaches.  With this approach, we were able to pretrain self-supervised video models on large-scale datasets. Because of the correspondence-aware pretraining, our models show zero-shot capabilities in \textit{any-point tracking}. Once fine-tuned, our models show very strong tracking performance across multiple datasets and metrics. In summary, in this paper, we proposed a self-supervised video pretraining approach, which exhibits strong tracking performance on zero-shot and after finetuning. 

{
    \small
    \bibliographystyle{ieeenat_fullname}
    \bibliography{references}
}

\clearpage
\appendix

\section{Supplementary Material}

This document provides a detailed description of the supplemental videos accompanying our paper, "Tracking by Predicting 3-D Gaussians Over Time." These videos offer qualitative insights into the performance of our proposed \approachName{} model, complementing the quantitative evaluations presented in the main manuscript. The videos, provided as MP4 files in the supplementary zip and referenced by filename in the sections below, showcase zero-shot tracking capabilities, tracking after fine-tuning, and visualizations of the pre-training reconstruction process. In addition to these videos, we provide full-page static figures that summarize representative frame sequences and qualitative comparisons between \approachName{} and prior work.

\subsection{Zero-shot Tracking Examples}

Our \approachName{} model is pre-trained through a self-supervised objective that enforces temporal correspondence by predicting the evolution of 3D Gaussian primitives over time. This pre-training allows the model to perform point tracking in a zero-shot manner, without any direct supervision from labeled tracking data. The following videos illustrate this emergent capability. This provides a qualitative complement to the zero-shot quantitative results reported in Table 2 of the main paper.

\begin{enumerate}
    \item \textbf{Zero-shot tracking on TAP-Vid Davis dataset:} Selected points are tracked across frames, showcasing the model's ability to maintain correspondence for objects undergoing motion and deformation in diverse visual contexts. These visualizations correspond to the supplementary video files \texttt{zeroshot-davis.mp4} and \texttt{zeroshot\_davis2.mp4}.
    \item \textbf{Zero-shot tracking on TAP-Vid Kinetics dataset:} Similar to the DAVIS examples, this video presents zero-shot tracking results on sequences from the Kinetics dataset. The scenes in Kinetics often feature a wide array of human actions and complex interactions, providing a different set of challenges. The video highlights the model's capacity to track points even in the presence of varied object classes and complex motion. These visualizations correspond to the supplementary video files \texttt{zeroshot-kinetics.mp4} and \texttt{zeroshot-kinetics2.mp4}.
    \item \textbf{Failure cases in zero-shot tracking:} To provide a balanced view of our model's zero-shot capabilities, this video illustrates common failure modes. As discussed in the main paper's Limitations section, \approachName{}'s zero-shot tracking performance can degrade under certain conditions. Specifically, scenes with complex, high-frequency backgrounds, combined with significant camera motion, pose a challenge. This is attributed to two main factors: (i) the pre-training assumption of a static camera, which is violated in such scenarios, and (ii) the fixed budget of 256 Gaussian primitives used during pre-training. In visually dense backgrounds, a substantial portion of these Gaussians may be allocated to modeling the static parts of the scene, leaving insufficient capacity to accurately represent the motion of foreground objects or to disambiguate object motion from camera-induced motion. This video exemplifies these situations where tracking accuracy diminishes. These visualizations correspond to the supplementary video file \texttt{zeroshot-failure.mp4}.
\end{enumerate}

\begin{figure}[t]
    \centering
    \includegraphics[width=\linewidth]{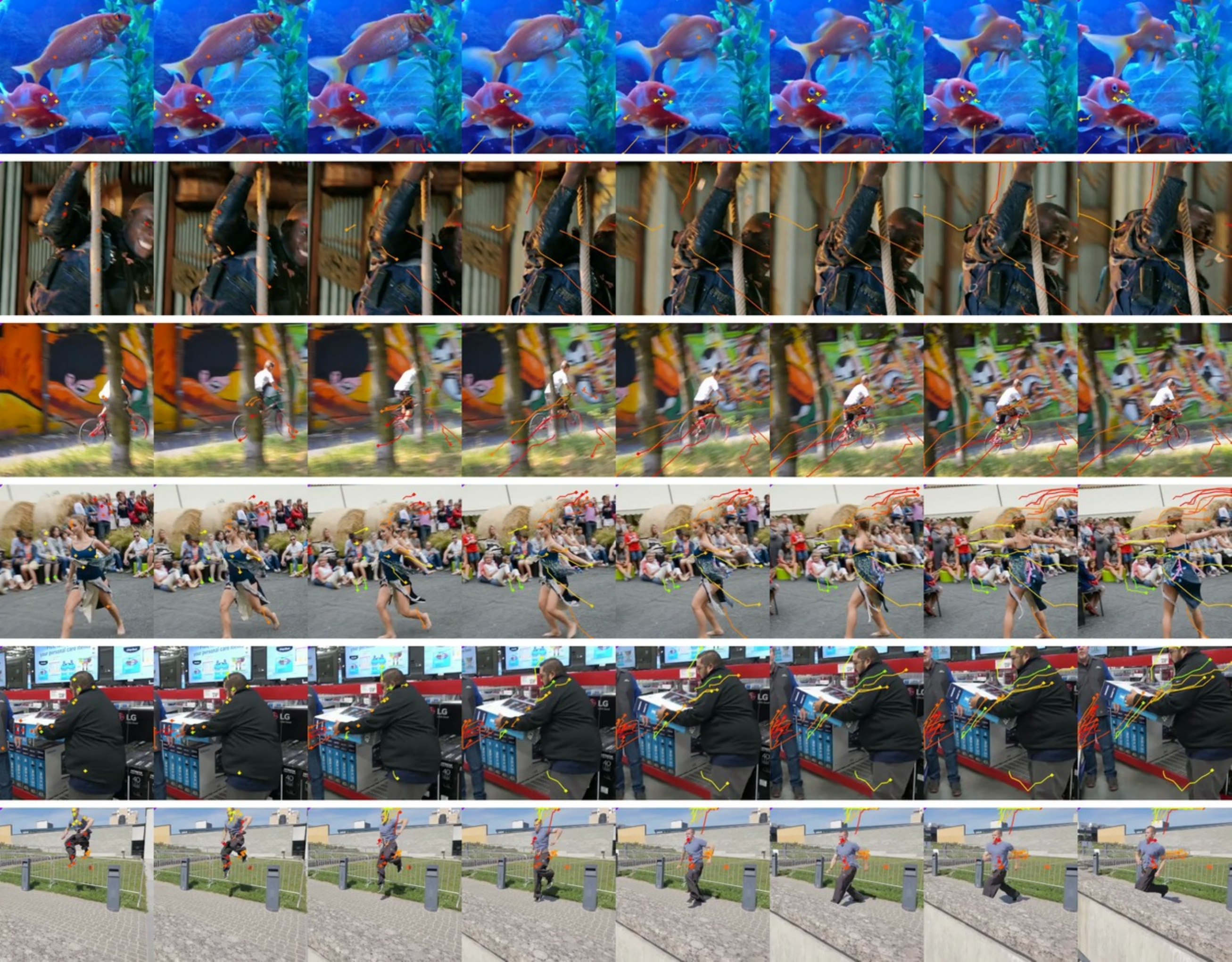}
    \caption{\textbf{Zero-shot failure cases.} Examples of challenging zero-shot tracking scenarios where \approachName{} degrades, typically due to complex, high-frequency backgrounds and substantial camera motion. The figure reflects failure modes discussed in the main paper, where the limited 256-Gaussian budget and the static-camera pre-training assumption can reduce tracking accuracy on moving foreground objects.}
    \label{fig:supp_zeroshot_failure}
\end{figure}

Figures~\ref{fig:supp_zeroshot_davis}, \ref{fig:supp_zeroshot_kinetics}, and \ref{fig:supp_zeroshot_failure} provide static visualizations corresponding to these zero-shot tracking examples. The figures highlight typical point trajectories on TAP-Vid DAVIS and TAP-Vid Kinetics, as well as characteristic failure modes in challenging settings, complementing the dynamic behavior shown in the videos.

\begin{figure}[t]
    \centering
    \includegraphics[width=\linewidth]{figures-pdf/supplemental-figures/zeroshot_davis.pdf}
    \caption{\textbf{Zero-shot tracking on TAP-Vid DAVIS.} Representative zero-shot point tracking results of \approachName{} on TAP-Vid DAVIS. Colored points denote tracked locations over time. The examples illustrate that our model can maintain long-range correspondences for deforming objects and complex motions without task-specific supervision.}
    \label{fig:supp_zeroshot_davis}
\end{figure}

\begin{figure}[t]
    \centering
    \includegraphics[width=\linewidth]{figures-pdf/supplemental-figures/zeroshot_kinetics.pdf}
    \caption{\textbf{Zero-shot tracking on TAP-Vid Kinetics.} Static visualization of zero-shot tracking with \approachName{} on TAP-Vid Kinetics sequences. Colored points denote tracked locations over time. The scenes involve diverse human actions and object interactions, and the model is able to track points across rapid motions and appearance changes.}
    \label{fig:supp_zeroshot_kinetics}
\end{figure}

\subsection{Qualitative Comparison with GMRW-C}

\begin{figure}[t]
    \centering
    \includegraphics[width=\linewidth]{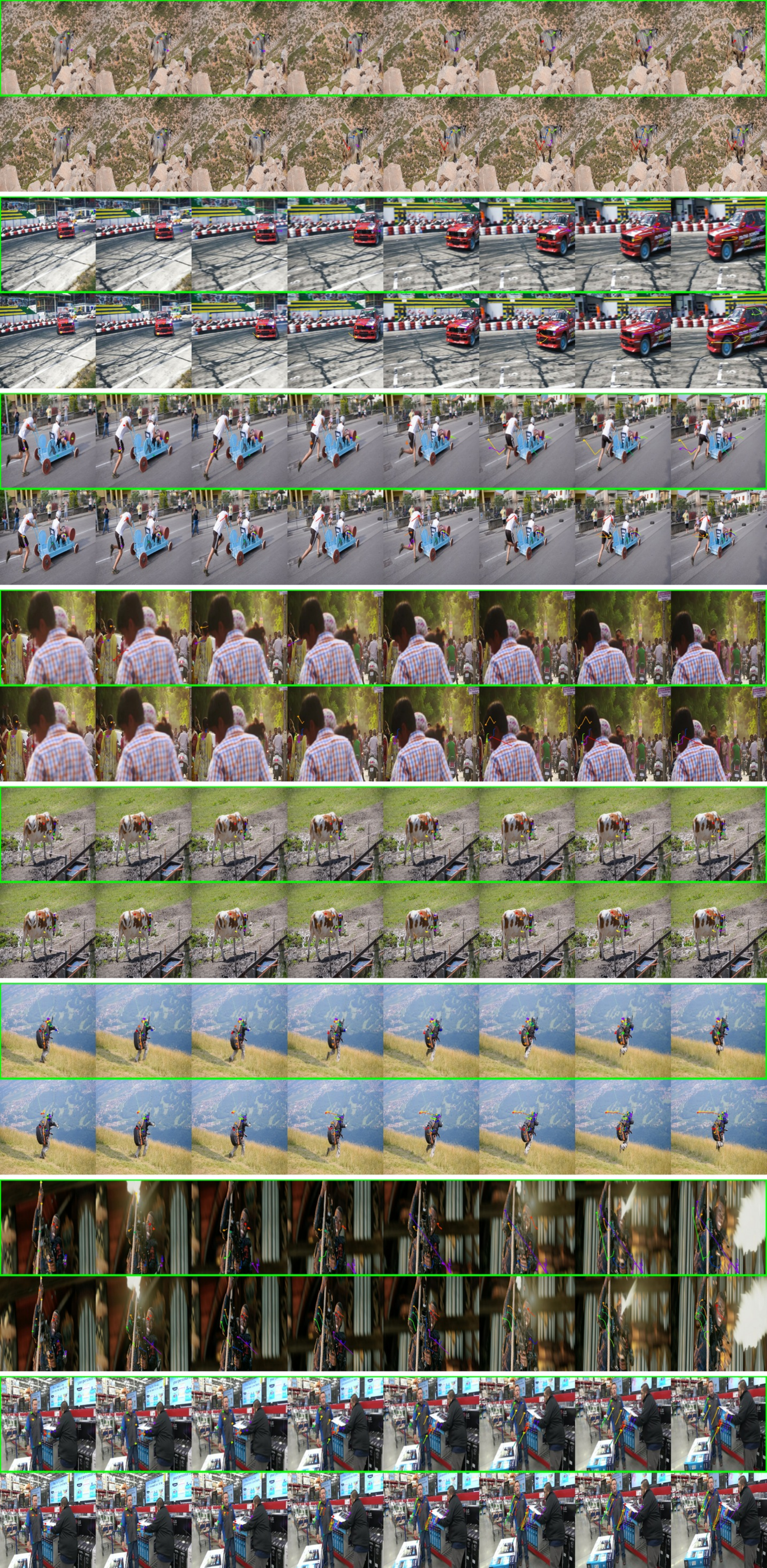}
    \caption{\textbf{Qualitative comparison with GMRW-C on TAP-Vid DAVIS.} Representative zero-shot tracking sequences comparing \approachName{}-zeroshot (the sequences with the green border) to GMRW-C~\cite{shrivastava2024gmrw} on TAP-Vid DAVIS. Each strip shows point trajectories over time for both methods on challenging motions and occlusions. On TAP-Vid Davis, \approachName{} is conservative at occluding tracks, but has a slightly worse tracking precision, which leads to a similar Average Jaccard score compared to GMRW-C.}
    \label{fig:supp_comparison_davis}
\end{figure}

\begin{figure}[t]
    \centering
    \includegraphics[width=\linewidth]{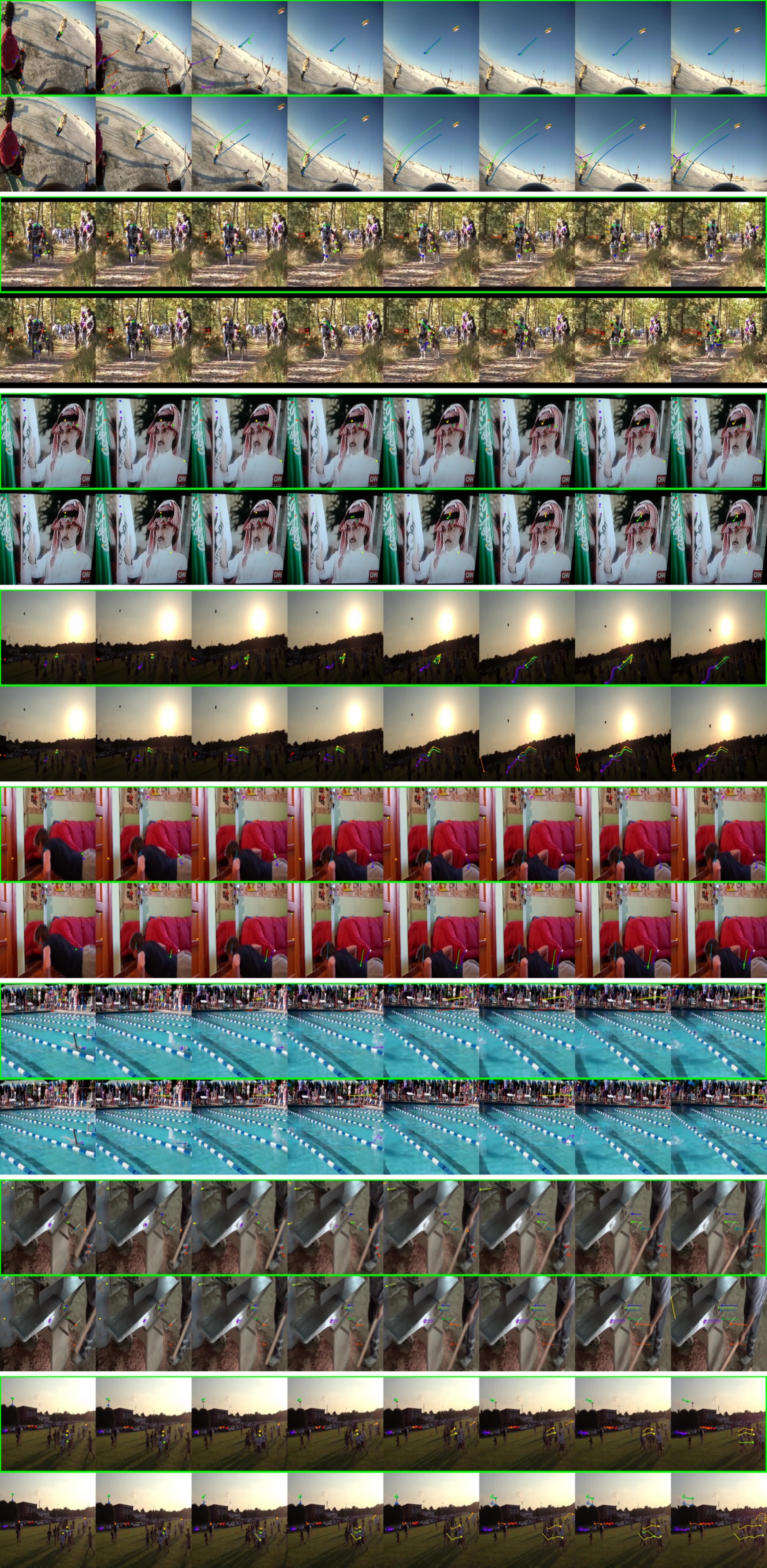}
    \caption{\textbf{Qualitative comparison with GMRW-C on TAP-Vid Kinetics.} Additional zero-shot tracking comparisons on TAP-Vid Kinetics sequences. While \approachName{}-zeroshot (the sequences with the green border) often yields smoother, more persistent tracks, GMRW-C can better preserve points on small, fine-scale structures undergoing large displacements. These examples illustrate the trade-offs between the two approaches and complement the quantitative metrics reported in the main paper.}
    \label{fig:supp_comparison_kinetics}
\end{figure}

The main paper reports quantitative zero-shot tracking metrics and presents a small set of qualitative comparisons between \approachName{}-zeroshot and the strongest self-supervised baseline, GMRW-C~\cite{shrivastava2024gmrw}. Here, we expand on these analyses by providing additional examples on TAP-Vid DAVIS and TAP-Vid Kinetics in Figures~\ref{fig:supp_comparison_davis} and \ref{fig:supp_comparison_kinetics}. The supplementary videos \texttt{comparison\_grid1.mp4} and \texttt{comparison\_grid2.mp4} provide dynamic visualizations of these qualitative comparisons.

Across many sequences, we observe that \approachName{}-zeroshot tends to produce temporally coherent tracks: once a point is visible, its identity is usually preserved without short-lived disappearances or rapid switches, leading to smoother trajectories over time. In contrast, GMRW-C sometimes exhibits frame-to-frame instabilities such as early occlusion predictions, brief drop-outs, or re-activations of points that reduce the effective temporal continuity of the track. We also see that our method often maintains point visibility until objects fully exit the frame, making it more conservative around occlusions, whereas GMRW-C may mark points as occluded several frames before they leave view and occasionally re-activate them spuriously.

At the same time, these comparison strips highlight scenarios where GMRW-C can be advantageous. In particular, when small, visually detailed regions undergo large displacements, GMRW-C sometimes preserves point locations more faithfully than \approachName{}-zeroshot. This is particularly evident for the TAP-Vid Davis videos, where \approachName{}-zeroshot predicts occlusions better but sacrifices point location precision, which is qualitatively consistent with the Average Jaccard score being similar between the two methods. This behavior is consistent with a limitation of our representation: modeling each scene with only 256 Gaussian primitives constrains the spatial resolution at which fine-scale details can be reconstructed and tracked. Overall, these additional comparisons reinforce the quantitative results and suggest that our Gaussian-based zero-shot tracker is well-suited for stable, occlusion-aware tracking, while leaving room for future work to improve fine-detail fidelity under large motions.

\subsection{Fine-tuned Tracking Examples}

While \approachName{} exhibits strong zero-shot tracking, its representations can be further adapted for enhanced performance via supervised fine-tuning on task-specific datasets. The following videos demonstrate the capabilities of our model after such fine-tuning, corresponding to the fine-tuned results presented in Table 2 of our main paper.

\begin{enumerate}
    \item \textbf{Fine-tuned tracking on TAP-Vid Davis dataset:} This video showcases the improved point tracking precision and robustness of \approachName{} on the TAP-Vid DAVIS dataset after supervised fine-tuning. The tracking is visibly more stable and accurate compared to the zero-shot results, especially in challenging sequences involving occlusions, rapid motion, and multi-object interactions. These sequences are provided in the supplementary video files \texttt{finetune-davis.mp4} and \texttt{finetune-davis2.mp4}.

    \item \textbf{Fine-tuned tracking on TAP-Vid Kinetics dataset:} Similar to the prior video, this video shows the improved point tracking on TAP-Vid Kinetics dataset after supervised fine-tuning. These sequences are provided in the supplementary video files \texttt{finetune-kinetics.mp4} and \texttt{finetune-kinetics2.mp4}.
\end{enumerate}

\begin{figure}[t]
    \centering
    \includegraphics[width=\linewidth]{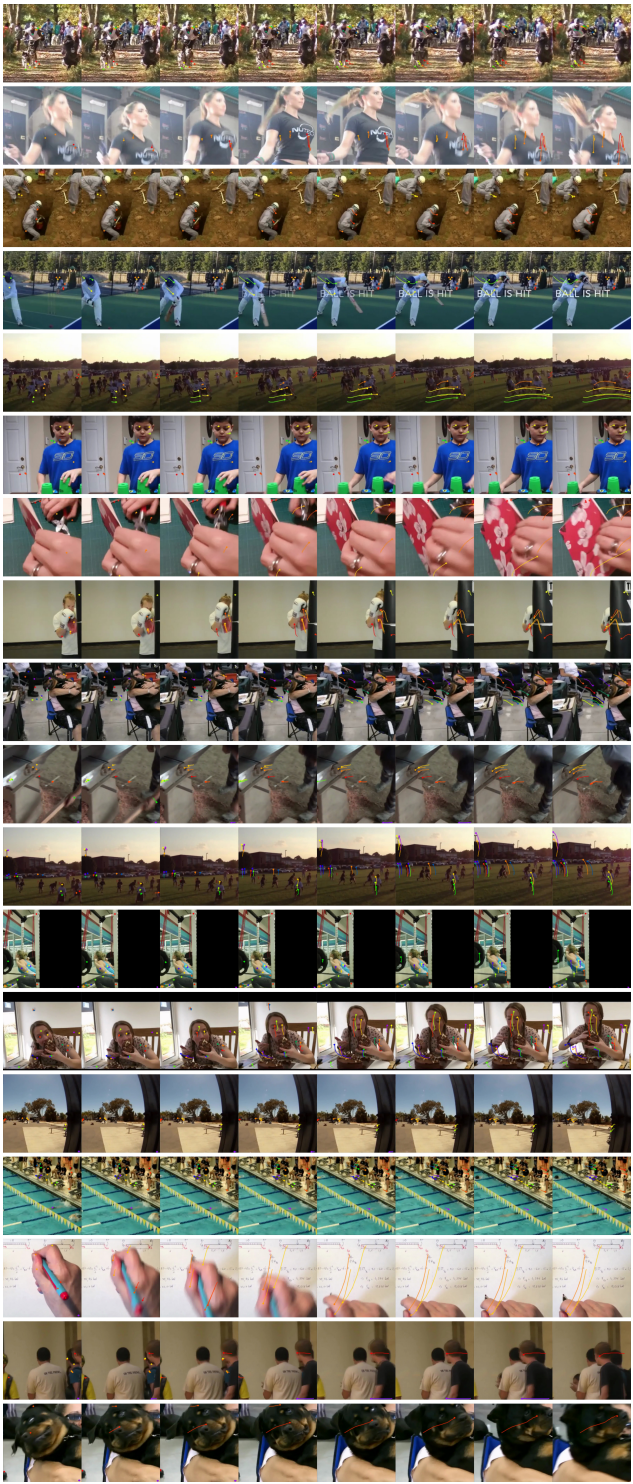}
    \caption{\textbf{Fine-tuned tracking on TAP-Vid Kinetics.} Static example of fine-tuned tracking on TAP-Vid Kinetics. Fine-tuning leads to crisper trajectories and more stable point identities across complex human actions, complementing the qualitative improvements observed in the corresponding videos.}
    \label{fig:supp_finetune_kinetics}
\end{figure}

Figures~\ref{fig:supp_finetune_davis} and \ref{fig:supp_finetune_kinetics} provide static examples of fine-tuned tracking on both TAP-Vid DAVIS and TAP-Vid Kinetics, illustrating the reduction in drift and jitter as well as improved robustness to occlusions compared to the zero-shot case.

\begin{figure}[t]
    \centering
    \includegraphics[width=\linewidth]{figures-pdf/supplemental-figures/finetune_davis.pdf}
    \caption{\textbf{Fine-tuned tracking on TAP-Vid DAVIS.} Still-frame visualization of point tracking with \approachName{} after supervised fine-tuning on TAP-Vid DAVIS. Compared to the zero-shot sequences in Figures~\ref{fig:supp_zeroshot_davis} and \ref{fig:supp_zeroshot_failure}, the fine-tuned model exhibits more precise localization, reduced temporal jitter, and improved handling of occlusions and object interactions.}
    \label{fig:supp_finetune_davis}
\end{figure}

\subsection{\approachName{} Pre-training Gaussians Rendered}

The core of \approachName{} lies in its self-supervised pre-training strategy, where the model learns to reconstruct video sequences by predicting the parameters and temporal dynamics of 3D Gaussian primitives within an MAE framework. The videos in this section show what the reconstructed videos look like.

\begin{enumerate}
    \item \textbf{Rendered Videos:} These three videos (\texttt{renders1.mp4}, \texttt{renders2.mp4}, and \texttt{renders3.mp4}) visualize the output of the \approachName{} decoder during the pre-training phase. For a given input video, the model predicts a set of 3D Gaussians for the initial frame and their subsequent transformations for the following frames. These Gaussians are then rendered to reconstruct the video sequence. The visualizations demonstrate the model's ability to represent scene content using a collection of dynamic Gaussian primitives.
\end{enumerate}

\begin{figure}[t]
    \centering
    \includegraphics[width=\linewidth]{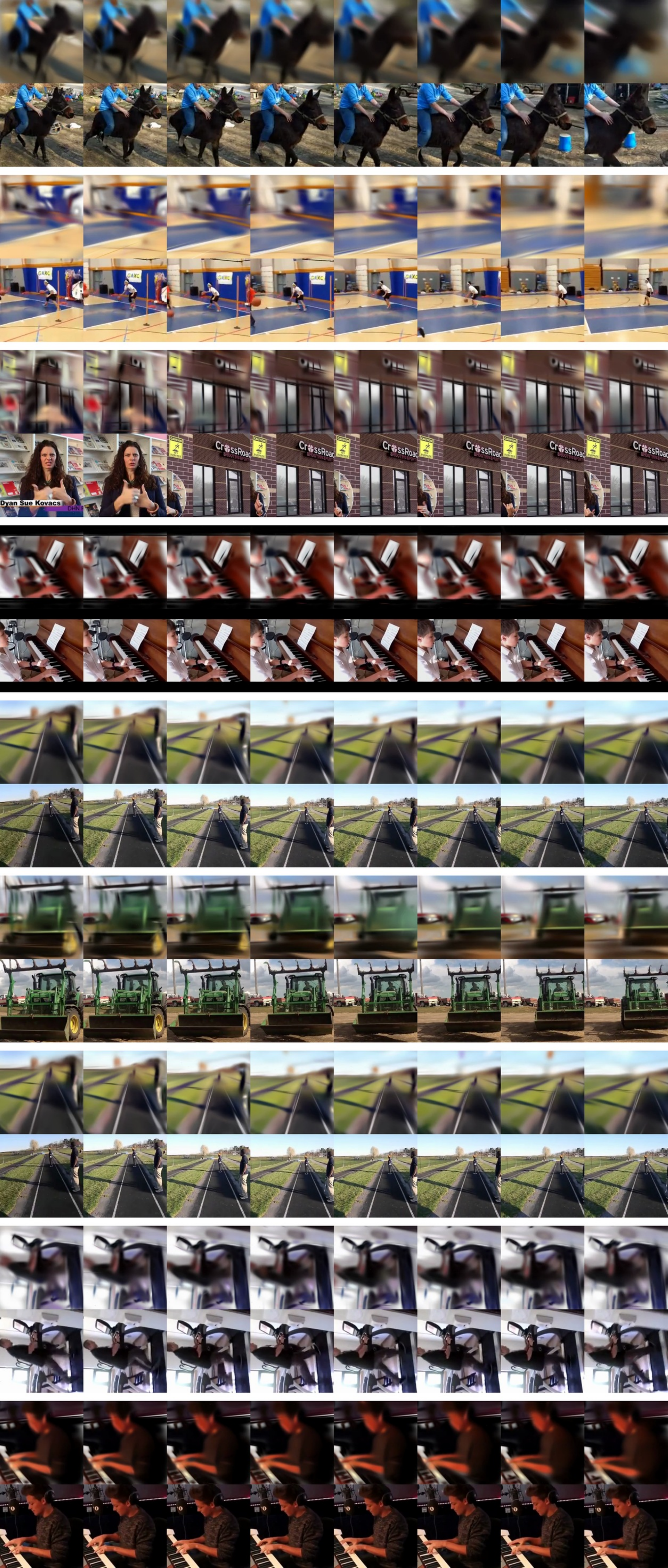}
    \caption{\textbf{\approachName{} pre-training reconstructions (set 1).} Example frames from the rendered outputs of the \approachName{} decoder during pre-training. For each sequence, the model predicts an initial set of 3D Gaussians and their temporal evolution, which are then rendered to reconstruct the video. The visualizations show that a compact set of dynamic Gaussian primitives can capture the dominant structure and motion in the scene.}
    \label{fig:supp_renders_part1}
\end{figure}

\begin{figure}[t]
    \centering
    \includegraphics[width=\linewidth]{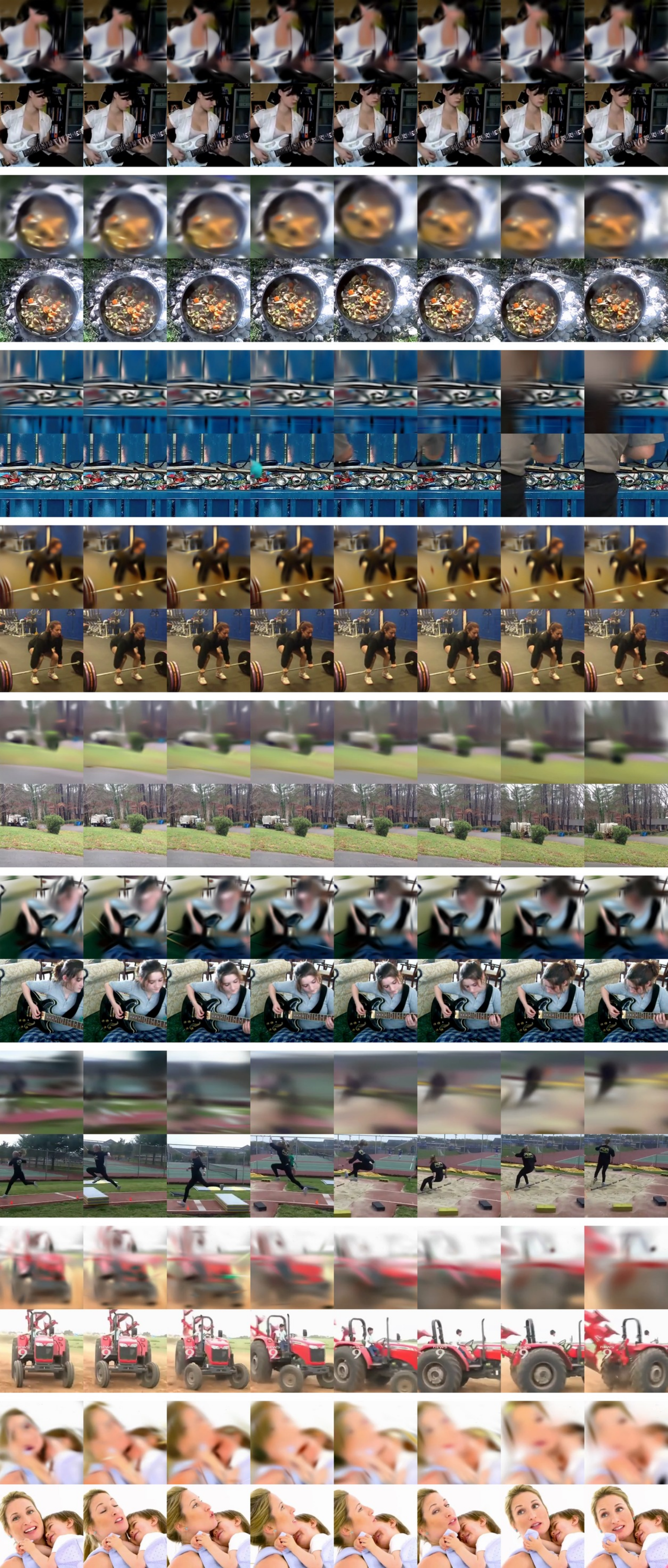}
    \caption{\textbf{\approachName{} pre-training reconstructions (set 2).} Additional examples of pre-training reconstructions on different sequences. While the Gaussians capture coarse geometry and dynamics, fine-scale textures and small structures are limited by the 256-Gaussian budget, consistent with the discussion of representation capacity in the main paper.}
    \label{fig:supp_renders_part2}
\end{figure}

Figures~\ref{fig:supp_renders_part1} and \ref{fig:supp_renders_part2} show static renderings from two sets of pre-training reconstructions. These stills highlight how a relatively small number of Gaussians can capture coarse geometry, appearance, and motion, while also revealing the limitations of the representation in highly textured or fine-detail regions.

\end{document}